\documentclass[twocolumn,10pt]{article} 

\usepackage{booktabs}
\usepackage{siunitx}
\usepackage{amsfonts}
\usepackage{natbib}
\usepackage{hyperref}
\usepackage{graphicx}
\usepackage{amsmath, amssymb, amsthm}
\usepackage{amsfonts}
\usepackage[margin=1in]{geometry}
\usepackage{natbib}
\usepackage{booktabs}
\usepackage{siunitx}
\usepackage{microtype}
\usepackage[noblocks]{authblk}

\usepackage[switch]{lineno}
\usepackage{silence}
\title{\textbf{Optimization Instability in Autonomous Agentic Workflows for Clinical Symptom Detection}}

\author[1]{Cameron Cagan}
\author[1]{Pedram Fard}
\author[1]{Jiazi Tian}
\author[1]{Jingya Cheng}
\author[2]{Shawn N. Murphy}
\author[1]{Hossein Estiri}

\affil[1]{Department of Medicine, Massachusetts General Hospital, Boston, MA, USA}
\affil[2]{Department of Neurology, Massachusetts General Hospital, Boston, MA, USA}

\date{}

\begin{document}

\maketitle

\begin{abstract}
Autonomous agentic workflows that iteratively refine their own behavior hold considerable promise, yet their failure modes remain poorly characterized. We investigate optimization instability, a phenomenon in which continued autonomous improvement paradoxically degrades classifier performance, using Pythia, an open-source framework for automated prompt optimization. Evaluating three clinical symptoms with varying prevalence (shortness of breath at 23\%, chest pain at 12\%, and Long COVID brain fog at 3\%), we observed that validation sensitivity oscillated between 1.0 and 0.0 across iterations, with severity inversely proportional to class prevalence. At 3\% prevalence, the system achieved 95\% accuracy while detecting zero positive cases, a failure mode obscured by standard evaluation metrics. We evaluated two interventions: a guiding agent that actively redirected optimization, amplifying overfitting rather than correcting it, and a selector agent that retrospectively identified the best-performing iteration successfully prevented catastrophic failure. With selector agent oversight, the system outperformed expert-curated lexicons on brain fog detection by 331\% (F1) and chest pain by 7\%, despite requiring only a single natural language term as input. These findings characterize a critical failure mode of autonomous AI systems and demonstrate that retrospective selection outperforms active intervention for stabilization in low-prevalence classification tasks.
\end{abstract}

\paragraph*{Data and Code Availability}
This paper uses a dataset of 400 chart-reviewed EHR notes, each annotated with binary yes/no classifications on various signs and symptoms by clinicians. The dataset is confidential and not publicly available. The code can be found \href{https://github.com/clai-group/Pythia}{here}.

\paragraph*{Institutional Review Board (IRB)}
This study was approved by the Mass General Brigham IRB.

\section{Introduction}
\label{sec:intro}

The effectiveness of large language models in specialized tasks depends critically on prompt engineering, the iterative process of crafting instructions that elicit desired behaviors from the model \citep{brown2020language, wei2022chain}. While prompt engineering has emerged as a fundamental skill for deploying LLMs in practical applications, it remains largely a manual craft that requires domain expertise, extensive trial-and-error experimentation, and implicit knowledge of model behavior. This manual approach introduces significant barriers: practitioners must invest substantial time testing variations, lack systematic frameworks for diagnosing failure modes, and often struggle to balance competing performance objectives such as precision and recall.

Current approaches to prompt optimization fall into two categories, each with fundamental limitations. Meta-learning methods that train models to generate effective prompts require extensive computational resources and large training datasets, making them inaccessible for many applications \citep{shin2020autoprompt, pryzant2023automatic}. Conversely, heuristic-based systems that apply rule-based transformations lack the semantic understanding necessary to address task-specific failure modes, often producing generic modifications that fail to capture the nuances of domain-specific requirements \citep{zhou2023large}. Recent work on autonomous agentic systems has demonstrated that multi-agent architectures can coordinate specialized reasoning within defined domains \citep{park2023generative}, suggesting an alternative paradigm in which LLMs themselves drive optimization through interpretable error analysis.

The challenge is particularly acute in binary classification tasks where performance depends on achieving precise thresholds for both sensitivity and specificity \citep{he2009learning}. Medical screening applications, content moderation systems, and fraud detection pipelines all require careful calibration of false-positive and false-negative rates. Manual prompt engineering for these applications becomes a complex optimization problem in which improving one metric often degrades another, and practitioners lack principled methods for navigating this trade-off space.

These challenges are amplified in clinical natural language processing, where the goal is often to detect signs and symptoms from unstructured clinical notes \citep{wang2018clinical, rajkomar2018scalable}. Unlike structured data fields, free text documentation exhibits enormous variability in how clinicians describe the same clinical phenomenon, using synonyms, abbreviations, hedging language, and implicit references that require contextual interpretation. Traditional approaches rely on expert-curated lexicons: manually assembled vocabularies of relevant terms that are time-intensive to develop, difficult to maintain, and often fail to capture the full linguistic diversity of clinical documentation \citep{lee2020biobert}. The problem is particularly acute for subjective symptoms like Long COVID brain fog, a debilitating cognitive syndrome affecting up to 64\% of Long COVID patients \citep{thaweethai2023pasc}. Brain fog encompasses heterogeneous presentations, including memory difficulties, poor concentration, word-finding problems, and slowed thinking, which patients struggle to articulate and clinicians document inconsistently \citep{cummings2023language, hampshire2024cognition}. Unlike shortness of breath or chest pain, which have established clinical vocabularies and validated screening instruments, brain fog lacks standardized terminology and computational detection methods \citep{davis2023longcovid, monje2022neurobiology}. With an estimated 65 million individuals worldwide affected by Long COVID, scalable methods for identifying brain fog from existing clinical documentation represent an urgent unmet need \citep{ceban2022fatigue}.

We present an empirical investigation of optimization instability in autonomous agentic workflows, using Pythia \citep{tian2026pythia}, an open-source framework that automates prompt optimization via interpretable error analysis and targeted refinement. Rather than treating prompts as opaque parameters to be tuned, Pythia implements a multi-agent system that diagnoses specific failure modes, synthesizes targeted improvements in natural language, and adaptively balances competing performance objectives. By executing all optimization operations on the target LLM itself, the framework ensures refinements are intrinsically compatible with the model's operational characteristics while maintaining full interpretability throughout the optimization process. Our central contribution is the characterization of optimization instability as a failure mode of autonomous systems operating under class imbalance, and the demonstration that retrospective selection provides more robust stabilization than active intervention.

\section{System Design and Architecture} 
\label{sec:system}

\subsection{Overview}

Pythia is an open-source framework for automated prompt optimization that systematically refines prompts to meet user-specified performance requirements. The system implements a multi-agent architecture in which specialized components collaborate to diagnose failures, generate targeted improvements, and perform self-validation. All optimization operations execute on the user-specified LLM, ensuring prompt development is inherently compatible with the model's characteristics.

The framework decomposes prompt engineering \citep{prompting} into distinct subtasks, each handled by dedicated agents. This modular design enables parallel analysis of different error types and maintains clear separation between evaluation and optimization \cite{nof1}.

\subsection{Agent Architecture}

The system comprises five specialized agents, each implementing a specific role in the optimization pipeline. Agents are constructed using structured LLM calls with engineered system prompts that define their behavior. The agents are implemented in Python and use LangChain \citep{langchain} for prompt formatting.

\subsubsection{Specialist Agent}

The Specialist Agent serves as the classifier within the optimization loop. Given prompt $P \in \mathcal{P}$, standard operating procedure $S$, and clinical note $x \in \mathcal{X}$, the agent generates a binary classification $\hat{y} = f_P(x) \in \{0, 1\}$. The agent's minimal design enables unbiased performance evaluation and clearly separates the optimization objective from the optimization mechanism. Performance metrics are computed by comparing predictions $\{\hat{y}_i\}_{i=1}^{n}$ against ground truth labels $\{y_i\}_{i=1}^{n}$, yielding sensitivity $\sigma = \text{TP}/(\text{TP} + \text{FN})$ and specificity $\tau = \text{TN}/(\text{TN} + \text{FP})$.

\subsubsection{Error Analysis Agents}

The Improver Agents employ counterfactual reasoning to diagnose errors, generating detailed natural-language explanations for each classification error.

The Specificity Improver analyzes false positives. For each instance $x \in \mathcal{FP} = \{x_i : \hat{y}_i = 1 \land y_i = 0\}$, it produces critique $c(x)$ identifying negative indicators that were overlooked. The Sensitivity Improver performs the analogous operation for false negatives, where for each $x \in \mathcal{FN} = \{x_i : \hat{y}_i = 0 \land y_i = 1\}$, it produces a critique $c(x)$ highlighting positive signals that were missed.

The key innovation is that, by generating natural-language explanations rather than abstract error metrics, Pythia produces semantically rich feedback that can be incorporated into subsequent prompt development. The Error Analysis Agents perform interpretable error attribution, converting classification failures into actionable insights expressible in the same representational space as the prompt itself.

\subsubsection{Synthesis Agents}

The Summarizer Agents implement multi-document synthesis that aggregates error critiques and generates targeted prompt modifications. Let $C = \{c_1, c_2, \ldots, c_m\}$ denote the set of critiques generated for errors of a particular type, $P_t$ the current prompt at iteration $t$, and $S$ the standard operating procedure. Each Summarizer generates refined prompt $P_{t+1} = g(C, P_t, S)$.

The Specificity Summarizer focuses on constraint tightening by identifying overly permissive criteria in $P_t$ and introducing exclusion rules and negative indicators. The Sensitivity Summarizer performs feature expansion by enriching $P_t$ with detection patterns that capture previously missed positive signals. The synthesis operates as a semantic differencing algorithm that identifies the gap between the current prompt's implicit decision boundary and the desired boundary, then generates natural language modifications to shift the boundary appropriately.

\subsection{Optimization Pipeline}

\subsubsection{Development Workflow}

The Development Workflow implements a multi-objective optimization protocol. Let $P_0$ denote the initial prompt, and let $\theta_\sigma$ and $\theta_\tau$ denote the sensitivity and specificity thresholds, respectively. At each iteration $t$, the system executes the following operations.

The system first performs evaluation and metric calculation by running the Specialist Agent on all instances in the development dataset $\mathcal{D}_{\text{dev}}$, computing predictions, and calculating performance metrics by comparing predictions to ground truth labels:
\begin{align}
\sigma_t &= \frac{|\{x : f_{P_t}(x) = 1 \land y(x) = 1\}|}{|\{x : y(x) = 1\}|}\\
\tau_t &= \frac{|\{x : f_{P_t}(x) = 0 \land y(x) = 0\}|}{|\{x : y(x) = 0\}|}
\end{align}

The system then performs convergence checking. If both $\sigma_t \geq \theta_\sigma$ and $\tau_t \geq \theta_\tau$, optimization terminates with $P^* = P_t$. Otherwise, the system prioritizes the metric that has not yet reached its threshold. When both metrics fall below the threshold, sensitivity receives priority by default.

Error analysis and prompt synthesis proceed by selecting the appropriate Improver Agent based on the target metric, generating critiques for all corresponding errors, filtering meaningful critiques, and invoking the corresponding Summarizer to generate $P_{t+1}$.

This protocol continues until convergence or until the maximum iteration count $T_{\max} = 7$ is reached. Dynamic prioritization enables the system to navigate the sensitivity specificity tradeoff, allocating computational resources to the limiting performance factor.

\subsubsection{Performance Degradation Prevention}

To mitigate potential negative performance impacts during iterative prompt development, Pythia includes a performance monitoring mechanism. After iteration $t > 0$, the system compares the current prompt's performance against the previous iteration. If $\text{F1}(P_t) < \text{F1}(P_{t-1})$, the system reverts to improving the previous prompt, providing the failed prompt as a negative example:
\begin{equation}
P_{t+1} = g(C, P_{t-1}, S, P_t^{\text{fail}})
\end{equation}

This mechanism serves two purposes: preventing optimization from spiraling toward suboptimal regions of the prompt space, and providing the Summarizer with explicit examples of modifications to avoid.

\subsubsection{Prompt Selection}

After completing all development iterations, Pythia selects the optimal prompt for validation based on performance on the development set. The system evaluates all generated prompts $\{P_0, P_1, \ldots, P_T\}$ and selects:
\begin{equation}
P^* = {P_t} \text{F1}_{\mathcal{D}_{\text{dev}}}(P_t)
\end{equation}

\subsubsection{Validation Workflow}

Following development, the Validation Workflow evaluates $P^*$ on the held out validation dataset $\mathcal{D}_{\text{val}}$. This serves two critical functions: confirming that improvements generalize beyond the development set, and detecting overspecialization to idiosyncrasies of $\mathcal{D}_{\text{dev}}$.

\subsubsection{Formal Characterization}

The optimization procedure can be formalized as iterative refinement in a discrete prompt space. Let $\mathcal{P}$ denote the space of valid prompts, $\mathcal{L}: \mathcal{P} \times \mathcal{D} \to \mathbb{R}$ a loss function (e.g., negative F1), and $\mathcal{T}: \mathcal{P} \times 2^{\mathcal{X}} \to \mathcal{P}$ the transformation operator implemented by the Summarizer Agents. The optimization seeks:
\begin{equation}
P^* = {P \in \{P_0, \mathcal{T}(P_0, \mathcal{E}_0), \ldots\}} \mathcal{L}(P, \mathcal{D}_{\text{val}})
\end{equation}
where $\mathcal{E}_t$ denotes the error set at iteration $t$. The key insight is that $\mathcal{T}$ operates in natural language space rather than continuous parameter space, enabling interpretable optimization but potentially introducing instabilities when the linguistic transformation fails to correspond to beneficial movement in the underlying decision boundary.

\begin{figure*}[!t]
\centering
\includegraphics[width=0.75\textwidth]{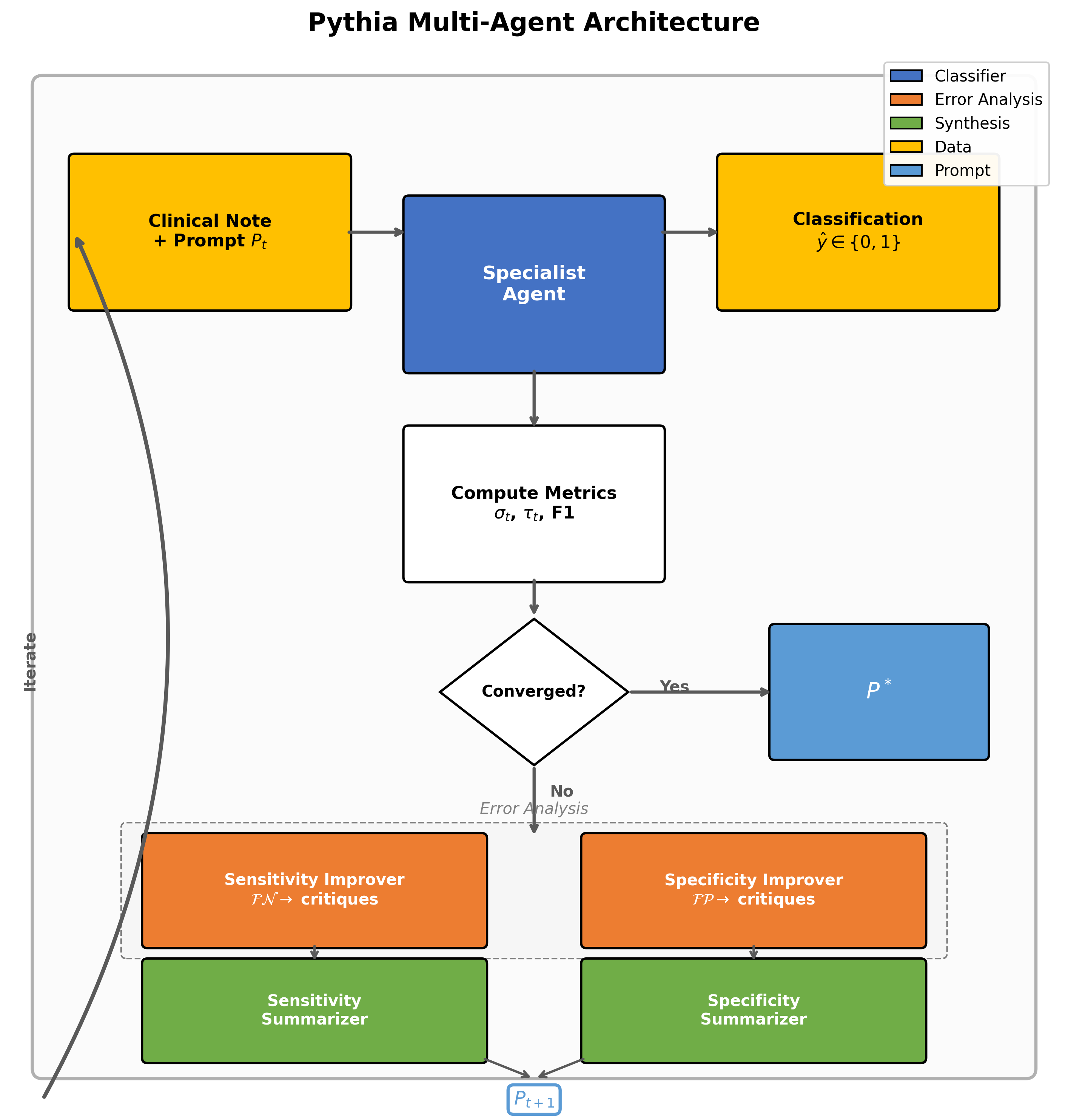}
\caption{\textbf{Pythia multi-agent architecture.} The optimization loop begins with a clinical note and prompt $P_t$ passed to the Specialist Agent, which produces binary classification $\hat{y} \in \{0,1\}$. The system computes sensitivity ($\sigma_t$), specificity ($\tau_t$), and F1 score. If the convergence criteria are met, the system outputs the final prompt $P^*$. Otherwise, Error Analysis Agents (orange) generate natural language critiques for false negatives ($\mathcal{FN}$) or false positives ($\mathcal{FP}$), which Synthesis Agents (green) aggregate into a refined prompt $P_{t+1}$. The loop iterates until convergence or reaches the maximum number of iterations.}
\label{fig:methodology}
\end{figure*}

\section{Methodology}
\label{sec:methods}

\subsection{Dataset and Task}

We evaluated Pythia on a dataset of 400 clinical notes from electronic health records, each annotated by clinicians with binary labels indicating the presence or absence of three symptoms: brain fog, chest pain, and shortness of breath (SOB). The dataset was partitioned into development ($n=200$) and validation ($n=200$) sets. Symptom prevalence varied substantially across conditions: SOB appeared in 23\% of notes, chest pain in 12\%, and brain fog in 3\%.

\subsection{Model Configuration}

All experiments used Llama 3.1 70B \citep{dubey2024llama} deployed locally via llama.cpp \citep{gerganov2023llamacpp}. We configured the model with a temperature of 0.0 and a maximum output length of 2048 tokens for deterministic classification, and prompt development.

\subsection{Baseline Comparison}

For each symptom, we constructed expert-curated lexicon baselines using clinical vocabulary terms identified through a systematic review of medical literature and clinical documentation. The lexicon classifier predicted positive when any lexicon term appeared in the clinical note.

\subsection{Study Design}

We evaluated the generalizability of Pythia, an autonomous agentic workflow originally developed for cognitive concern detection \citep{tian2026pythia}, to three clinical symptoms from an independent dataset: shortness of breath (23\% prevalence), chest pain (12\%), and brain fog (3\%). Unlike the original study, which utilized multiple clinical notes per patient, this evaluation used a single note per patient to assess performance under more constrained conditions.

\subsection{Experimental Protocol}

We initialized Pythia with minimal prompts containing only the target symptom name (e.g., ``brain fog''), providing no additional clinical context or detection strategies. This neutral initialization maximizes the observable improvement from autonomous optimization while providing a fair baseline across tasks. We ran seven optimization iterations per symptom and recorded performance metrics at each iteration on both development and validation sets.

\section{Results}
\label{sec:results}

\subsection{Autonomous Optimization Can Spiral}

Across seven optimization iterations, we observed prevalence-dependent instability in which continued optimization degraded rather than improved performance. Table~\ref{tab:spiral} presents validation performance at the first and final iterations.

\begin{table*}[t]
\caption{Validation performance degradation from first to final iteration. Continued autonomous optimization reduced F1 scores across all conditions, with severity inversely proportional to class prevalence.}
\label{tab:spiral}
\centering
\begin{tabular}{lcccc}
\toprule
Condition & Prevalence & Iter 1 F1 & Iter 7 F1 & $\Delta$ \\
\midrule
Brain fog & 3\% & 0.25 & 0.09 & $-$64\% \\
Chest pain & 12\% & 0.72 & 0.60 & $-$17\% \\
SOB & 23\% & 0.75 & 0.41 & $-$45\% \\
\bottomrule
\end{tabular}
\end{table*}

Brain fog exhibited the most severe instability. Validation sensitivity oscillated between 0.0 and 1.0 across iterations, reaching complete collapse (zero sensitivity) at iterations 3 and 6 (Figure~\ref{fig:spiral}A). At these points, the system detected no positive cases while maintaining high accuracy (0.95 to 0.97) due to class imbalance, a failure mode masked by standard evaluation metrics.

The spiral followed a characteristic pattern: the system attempted to improve specificity; this overcorrection reduced sensitivity; attempts to recover sensitivity destabilized the system further; and the cycle repeated with increasing amplitude. Higher-prevalence conditions exhibited attenuated oscillations but still showed performance degradation by the final iteration.

\subsection{Intervention 1: Guiding Agent}

To address the optimization spiral, we introduced a guiding agent that monitored performance after each iteration. When results failed to improve over the previous iteration, the agent paused optimization and instructed the system to abandon the current path and explore an alternative optimization direction.

\begin{table*}[t]
\caption{Guiding agent performance. The intervention reduced oscillation on the development set but amplified overfitting, as evidenced by the gap between development and validation performance.}
\label{tab:guiding}
\centering
\begin{tabular}{lccc}
\toprule
Condition & Dev F1 & Val F1 & Gap \\
\midrule
Brain fog & 0.47 & 0.00 & 0.47 \\
Chest pain & 0.73 & 0.64 & 0.09 \\
SOB & 0.70 & 0.75 & $-$0.05 \\
\bottomrule
\end{tabular}
\end{table*}

Table~\ref{tab:guiding} presents the best performance achieved with the guiding agent. The intervention successfully reduced oscillation on the development set; however, validation performance did not improve correspondingly. For brain fog, the guiding agent achieved the highest development F1 (0.47), while the validation F1 collapsed to 0.0. The intervention exacerbated overfitting by learning to optimize patterns in the development set that failed to generalize. The agent effectively learned to exploit the development set more aggressively rather than learning generalizable features.

\subsection{Intervention 2: Selector Agent}

We implemented an alternative approach: a selector agent that allowed all 7 optimization iterations to complete, then retrospectively selected the iteration with the best development-set F1 score for application to the validation set. This approach prioritized preventing catastrophic failure over maximizing performance.

\begin{table}[t]
\caption{Selector agent compared to expert curated lexicon baselines on the validation set. Pythia outperformed lexicons on brain fog (+331\%) and chest pain (+7\%) while achieving comparable performance on SOB ($-$5\%).}
\label{tab:selector}
\centering
\setlength{\tabcolsep}{3pt}
\begin{tabular}{lcccc}
\toprule
Condition & Prev. & Pythia F1 & Lexicon F1 & $\Delta$ \\
\midrule
Brain fog & 3\% & 0.25 & 0.058 & +331\% \\
Chest pain & 12\% & 0.72 & 0.67 & +7\% \\
SOB & 23\% & 0.75 & 0.79 & $-$5\% \\
\bottomrule
\end{tabular}
\end{table}

Table~\ref{tab:selector} compares the selector agent against expert curated lexicon baselines. The selector agent prevented catastrophic failure and outperformed lexicons on two of three conditions. For brain fog, the lexicon achieved perfect sensitivity (1.00) but zero specificity (0.00), flagging all cases as positive, yielding an F1 score of 0.058. Pythia achieved superior discrimination (F1 = 0.25) despite the 3\% prevalence. Notably, Pythia required only a single natural language term as input, whereas lexicons required expert curation of domain-specific vocabularies.

\begin{figure*}[t]
\centering
\includegraphics[width=\textwidth]{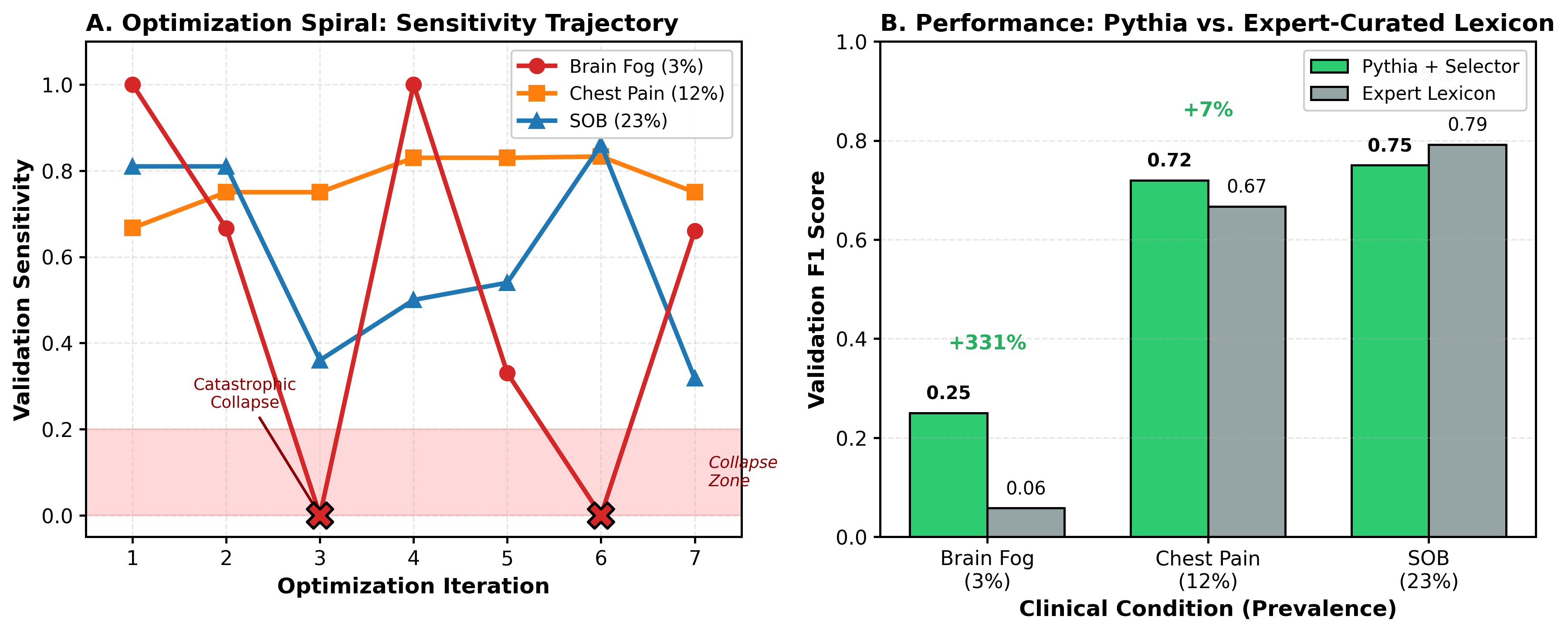}
\caption{\textbf{Optimization dynamics and comparative performance.} \textbf{(A)} Validation sensitivity across seven optimization iterations. Brain fog (3\% prevalence) exhibited catastrophic collapse, reaching zero sensitivity at iterations 3 and 6 (marked with X). Higher prevalence conditions showed oscillation but avoided complete failure. \textbf{(B)} Validation F1 scores comparing Pythia with selector agent versus expert curated lexicon baselines. Pythia outperformed lexicons on brain fog (+331\%) and chest pain (+7\%), while achieving comparable performance on SOB ($-$5\%), despite requiring only a single natural-language term as input.}
\label{fig:spiral}
\end{figure*}

Figure~\ref{fig:spiral}B illustrates the comparative performance. Across conditions, Pythia with selector agent oversight matched or exceeded lexicon performance while requiring minimal human input.

\subsection{Selection Was Conservative But Not Optimal}

The selector agent's conservative strategy of choosing the best development iteration prevented catastrophic failure but did not always identify the iteration with optimal generalization. Post hoc analysis revealed that alternative iterations would have outperformed the selected iteration for some conditions (Table~\ref{tab:optimal}).

\begin{table}[t]
\caption{Selected versus optimal iteration on the validation set. Development set performance did not reliably predict validation generalization, particularly at low prevalence.}
\label{tab:optimal}
\centering
\begin{tabular}{lcccc}
\toprule
Condition & Selected & Val F1 & Optimal & Val F1 \\
\midrule
Brain fog & Iter 1 & 0.25 & Iter 2 & 0.44 \\
Chest pain & Iter 1 & 0.72 & Iter 1 & 0.72 \\
SOB & Iter 1 & 0.75 & Iter 6 & 0.76 \\
\bottomrule
\end{tabular}
\end{table}

For brain fog, iteration 2 would have achieved F1 = 0.44 compared to the selected iteration's F1 = 0.25, a 76\% improvement. This gap highlights a fundamental challenge: at low prevalence, development set performance is an unreliable proxy for generalization because each iteration is informed by few positive cases.

\section{Discussion}
\label{sec:discussion}

This study demonstrates that Pythia, originally developed for cognitive concern detection, generalizes effectively to diverse clinical symptoms, including shortness of breath, chest pain, and Long COVID brain fog. Across all three conditions, Pythia with selector-agent oversight matched or exceeded the performance of expert-curated lexicons while requiring only a single natural-language term as input, rather than extensive domain-knowledge engineering. This finding has significant practical implications, as traditional lexicon-based approaches require clinical experts to enumerate comprehensive vocabularies of relevant terms, a process that is time-intensive, domain-specific, and difficult to scale. Pythia circumvents this bottleneck by autonomously learning discriminative patterns from minimal semantic input, dramatically lowering the barrier to clinical NLP applications.

Brain fog represents one of the most clinically challenging symptoms to detect computationally. Unlike shortness of breath or chest pain, which have established clinical vocabularies and diagnostic criteria, brain fog is characterized by subjective, heterogeneous presentations, including memory difficulties, poor concentration, word-finding problems, and cognitive fatigue \citep{cummings2023language}. Patients often struggle to articulate these experiences, and clinicians lack standardized screening instruments \citep{hampshire2024cognition}. The RECOVER initiative identified brain fog as one of the most prevalent Long COVID symptoms, affecting up to 64\% of patients in some cohorts \citep{thaweethai2023pasc}, yet detection in clinical notes remains exceptionally difficult because the linguistic markers are subtle, variable, and often inconsistently documented. No blood test, imaging study, or validated NLP tool currently exists for systematic brain fog screening \citep{davis2023longcovid}. Against this backdrop, Pythia's performance on brain fog detection merits careful interpretation. At 3\% prevalence, reflecting the proportion of patients with documented brain fog in our dataset, the system achieved F1 = 0.25, substantially outperforming the expert-curated lexicon (F1 = 0.058). The lexicon's failure mode was instructive: it achieved perfect sensitivity by flagging every case as positive, providing no discrimination whatsoever. Pythia, starting from only the phrase ``brain fog,'' learned to identify cases with meaningful specificity. While F1 = 0.25 may appear modest in absolute terms, it represents a 331\% improvement over the curated baseline under extraordinarily challenging conditions: single clinical notes, 3\% prevalence, and minimal semantic input. Post hoc analysis revealed that an alternative iteration would have achieved F1 = 0.44, a performance level that approaches clinical utility for a symptom with no existing computational detection method.

A central contribution of this work is the characterization of optimization instability in autonomous agentic systems. We observed that continued self-optimization can degrade performance, causing the system to oscillate between overcorrection and collapse. This instability was prevalence dependent, more pronounced at 3\% (brain fog) than at 23\% (SOB), suggesting that a sparse positive signal amplifies optimization noise. The phenomenon can be understood through the lens of empirical risk minimization under class imbalance \citep{he2009learning}: when positive examples are rare, small perturbations to the decision boundary can dramatically alter sensitivity while having minimal impact on overall accuracy, creating a loss landscape in which the optimization trajectory becomes unstable. We evaluated two interventions with divergent outcomes. The guiding agent, which actively redirected optimization upon detecting stagnation, paradoxically amplified overfitting; the system learned to exploit patterns in the development set more aggressively without improving generalization. The selector agent, which passively identified the best iteration post hoc, successfully prevented catastrophic failure while preserving competitive performance. This finding carries implications beyond clinical NLP. As autonomous AI systems increasingly engage in self-optimization during deployment, understanding when and why such optimization fails becomes critical. Our results suggest that active intervention during optimization may be counterproductive, whereas retrospective selection from a trajectory of attempts may be safer and more effective. The underlying intuition is that active intervention introduces additional degrees of freedom that the system can exploit to overfit, whereas retrospective selection constrains the search to prompts generated without knowledge of the selection criterion.

Several limitations warrant discussion. First, our evaluation used single clinical notes per patient, whereas real-world deployment might leverage longitudinal documentation. Future work should evaluate Pythia on multi-note patient records, where a cumulative signal may improve detection, particularly for subtle symptoms such as brain fog. Second, Pythia currently operates on minimal semantic input: a single natural language term. While this design choice maximizes accessibility, it may constrain performance on complex symptoms. We hypothesize that augmenting Pythia with richer initialization strategies, such as embedding-based initialization using clinical word embeddings trained on large corpora, integration of publicly available medical vocabularies from UMLS or SNOMED CT, or web search augmentation that retrieves current clinical literature on the target symptom, could substantially improve results. Third, our selector agent used the development F1 as its selection criterion. More sophisticated approaches, such as validation-aware selection, ensemble methods, or learned stopping criteria, may improve the correspondence between selected and optimal iterations. Fourth, our dataset reflects a single institution's documentation practices, and multi-site validation is necessary to establish generalizability across diverse clinical environments and documentation styles.

Pythia provides a powerful, generalizable tool for detecting clinical symptoms from unstructured notes. Despite minimal input requirements, the system matched or exceeded the performance of expert-curated lexicons across three diverse clinical conditions. For Long COVID brain fog, a symptom with no existing computational detection method, Pythia achieved over 300\% improvement relative to lexicon baselines under challenging low-prevalence conditions. We characterized optimization instability as a failure mode of autonomous agentic systems and demonstrated that retrospective selection outperforms active intervention in mitigating this instability. These findings establish Pythia as a practical tool for clinical NLP and identify clear directions for improvement. As healthcare systems increasingly seek scalable methods for symptom surveillance and phenotyping, autonomous agentic workflows offer a promising path forward, provided their failure modes are carefully understood and mitigated.

\bibliographystyle{plainnat}
\bibliography{citations}

\appendix

\end{document}